# Iris R-CNN: Accurate Iris Segmentation in Non-cooperative Environment

Chunyang Feng    Yufeng Sun    Xin Li

Duke Kunshan University

**Abstract**

*Despite the significant advances in iris segmentation, accomplishing accurate iris segmentation in non-cooperative environment remains a grand challenge. In this paper, we present a deep learning framework, referred to as Iris R-CNN, to offer superior accuracy for iris segmentation. The proposed framework is derived from Mask R-CNN, and several novel techniques are proposed to carefully explore the unique characteristics of iris. First, we propose two novel networks: (i) Double-Circle Region Proposal Network (DC-RPN), and (ii) Double-Circle Classification and Regression Network (DC-CRN) to take into account the iris and pupil circles to maximize the accuracy for iris segmentation. Second, we propose a novel normalization scheme for Regions of Interest (RoIs) to facilitate a radically new pooling operation over a double-circle region. Experimental results on two challenging iris databases, UBIRIS.v2 and MICHE, demonstrate the superior accuracy of the proposed approach over other state-of-the-art methods.*

## 1. Introduction

Iris recognition is one of the important approaches for the human recognition and authentication. Since the seminal work proposed by Daugman in 1993 [1], remarkable advances have been achieved for iris recognition based upon images with near infrared (NIR) illumination under controlled environment [2]. However, iris recognition in unconstrained environment remains a grand challenge today. In such circumstances, iris images often comprise significant adverse distortions due to off-axis, blur, occlusions, specular highlights and noise.

A complete iris recognition system must accomplish two major sub-tasks: iris segmentation and iris recognition. Iris segmentation aims to detect the inner and outer boundaries of an iris region and meanwhile generate an iris mask to distinguish the iris and non-iris pixels. The accuracy of iris recognition in unconstrained environment is highly limited by the quality of iris segmentation [3][4][5]. Even a small segmentation error may substantially degrade the accuracy of iris recognition [6]. Therefore, accurate iris segmentation is of great importance when designing an iris recognition system in unconstrained environment.

In the literature, a variety of novel methods have been proposed for iris segmentation, including integro-differential operator [7], unsupervised learning [8], improved Hough transform [9], etc. In addition to these traditional approaches that rely on handcrafted features and dedicated pre- and post-processing steps, modern approaches based on deep learning have recently gained popularity and achieved the state-of-the-art accuracy [10][11][12][13]. These approaches often use deep networks for pixel-wise semantic segmentation to distinguish iris and non-iris regions. However, they do not explicitly identify the iris and pupil circles that are extremely important for iris normalization and, consequently, the overall accuracy of iris recognition.

Unlike the aforementioned approaches focusing on semantic segmentation, we fundamentally re-think the conventional wisdom and cast iris segmentation into an instance segmentation problem. Based upon Mask R-CNN [14] that has been demonstrated with superior accuracy for instance segmentation, we derive Iris R-CNN to offer superior accuracy on iris segmentation over other state-of-the-art methods in non-cooperative environment. Iris R-CNN generates segmented and normalized iris images and masks, providing the required inputs for iris recognition.

Iris R-CNN does not simply re-implement the existing algorithms from Mask R-CNN. Instead, a number of core algorithms are re-designed by carefully exploring the unique characteristics of iris. First, Iris R-CNN takes into account the iris and pupil circles to maximize the accuracy for iris segmentation. Towards this goal, we propose two novel networks: (i) Double-Circle Region Proposal Network (DC-RPN), and (ii) Double-Circle Classification and Regression Network (DC-CRN). In contrast to the conventional RPN that captures a number of rectangular Regions of Interest (RoIs) only, DC-RPN uses a set of double-circle anchors to generate the RoIs where each RoI is specified by two non-concentric circles, corresponding to the iris and pupil circles respectively. A new double-circle regression scheme is developed to accurately determine the center coordinates and radii of both circles. DC-CRN further applies the aforementioned double-circle idea to ROI features to accurately classify and locate the iris region of interest.

Second, we propose a novel normalization scheme for RoI to replace the RoIAlign method of Mask R-CNN. It is based on the rubber sheet model previously used for iris normalization [3]. It remaps each pixel within RoI from the original Cartesian coordinate system to the polar coordinate system defined by the iris and pupil circles. The radial coordinate of the projected polar system ranges from the inner pupil boundary to the outer iris boundary. Compared to the conventional RoI pooling scheme that is often applied to a rectangular region, our proposed approach facilitates a radically new RoI pooling operation over a double-circle region. The proposed normalization scheme produces a set of normalized iris images and masks that are required by the subsequent steps for iris recognition.

The remainder of this paper is organized as follows. In Section 2, we brief review the literature on iris segmentation and recognition. Our proposed Iris R-CNN is developed in Section 3, and the corresponding iris recognition system based on Iris R-CNN is described in Section 4. The efficacy of Iris R-CNN is demonstrated by the experimental results with several public datasets in Section 5. Finally, we conclude in Section 6.

## 2. Related Works

In this section, we briefly review the relevant works for iris segmentation, iris recognition and Mask R-CNN.

**Iris segmentation:** The objective of iris segmentation is to detect the inner and outer boundaries of an iris region and meanwhile generate an iris mask to distinguish the iris and non-iris pixels. The integro-differential operator is one of the classical methods that have been widely adopted by commercial and open-source systems for iris segmentation [3]. The operator behaves as a circular edge detector. Namely, it detects iris circular edges by iteratively searching for a maximum response of an integro-differential expression.

To improve the accuracy of the integro-differential operator, a robust segmentation algorithm has been proposed to first approximately cluster the iris and non-iris regions and then use an enhanced integro-differential operator to accurately locate the iris and pupil boundaries under unconstrained environments [7]. An unsupervised method is developed in [8] by employing a polynomial upper eyelid model and a fully multispectral spatial Markovian texture model. Furthermore, a novel total-variation-based formulation is derived in [9] to robustly suppress noisy texture pixels for accurate iris localization.

Recently, deep neural networks have been adopted for accurate iris segmentation. Liu et al. apply Fully Convolutional Networks (FCNs) to semantic segmentation [10]. Afterwards, several similar semantic segmentation methods have been proposed. For instance, Arsalan et al. combine DenseNet and SegNet for iris segmentation [11]. Bazrafkan et al. use four newly designed FCNs in parallel for end-to-end segmentation [12]. Arsalan et al. adopt the VGG-Face network for iris images [13]. Although these approaches can accurately produce iris masks, they do not explicitly detect the iris and pupil circles. More recently, Zhao et al. propose a hybrid approach that uses a total variation model [9] to localize iris and pupil circles and applies FCN to normalized iris images for masking [18].

**Iris recognition:** Most classical methods for iris recognition often rely on carefully-designed handcrafted features, such as Gabor filter [3], LBP [15] and ordinal measures [16]. A growing body of literature on deep iris recognition has emerged recently. DeepIrisNet applies Convolutional Neural Networks (CNNs) supervised by a softmax loss function to extract iris features [17]. Zhao et al. propose to use FCNs to extract iris features encompassing different levels of details [18]. They introduce an extended triplet loss function to incorporate bit shifting and iris masking when learning discriminative iris features. Although these state-of-the-art approaches can achieve superior recognition accuracy, they rely on conventional segmentation methods that often become the bottleneck limiting the overall accuracy of an iris recognition system.

**Mask R-CNN:** The Mask R-CNN approach [14] efficiently detects objects and simultaneously generates a segmentation mask for each instance. It consists of three major steps. First, a CNN backbone architecture is used to extract feature maps. Second, a RPN takes these features to produce instance proposals, in the form of coordinates of bounding boxes. The bounding box proposals are used as the input to an RoIAlign layer, which interpolates the features in each bounding box to extract a fixed-sized representation. Third, the features of each RoI are passed to several detection branches, producing refined bounding box coordinates, a class prediction, and a binary mask for the predicted class. Mask R-CNN is the state-of-the-art method for instance segmentation, and has shown promising results in various applications such as pose estimation [19] and video prediction [20].

## 3. Iris R-CNN

In this section, we develop our proposed deep learning framework, Iris R-CNN, for iris segmentation. Iris R-CNN is derived from Mask R-CNN. It takes an iris image as the input and produces the normalized iris image and the corresponding mask, as shown in Figure 1. Similar to Mask R-CNN, Iris R-CNN consists of several major components: (i) a CNN backbone network for feature extraction, (ii) a DC-RPN, (iii) a DC-CRN for iris detection, and (iv) a mask prediction network.

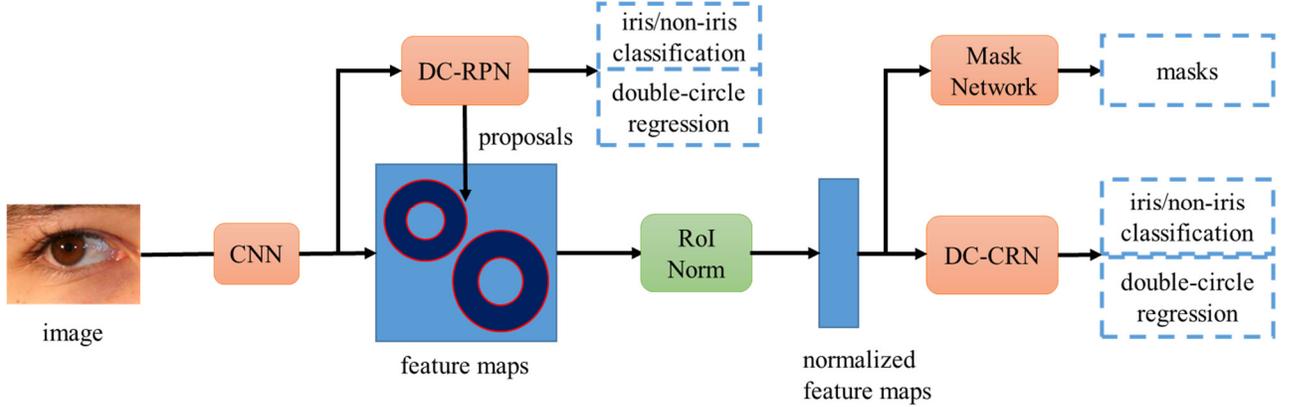

Figure 1: The proposed Iris R-CNN consists of (i) a CNN backbone network, (ii) a Double-Circle Region Proposal Network (DC-RPN), (iii) a Double-Circle Classification and Regression Network (DC-CRN), and (iv) a mask prediction network.

Following the same practice as Mask R-CNN, we choose the ResNet-101 Feature Pyramid Network (FPN) as our backbone. The DC-RPN is a lightweight FCN with 512 3×3 convolutional layers followed by two sibling 1×1 convolutional layers, which are used for iris/non-iris classification and double-circle regression respectively. The double-circle proposals produced by the DC-RPN are then used as the input to an RoI normalization layer, which interpolates the features in each double-circle region to extract a fixed-sized representation.

The DC-CRN is composed of two hidden 1,024-dimensional Fully-Connected (FC) layers followed by two sibling FC layers for the final classification and double-circle regression respectively. The mask prediction network is a FCN, composed of four 3×3 convolutional layers followed by a 2×2 transposed convolutional layers with stride 2, and a final 1×1 convolutional layer outputting the final binary mask indicating the iris/non-iris pixel at each spatial location.

In the following subsections, we will describe the details on these major components and highlight their novelties, including DC-RPN, RoI normalization operation, DC-CRN and mask generation. Seamlessly integrating these components yields a segmentation model that can achieve superior accuracy on extremely challenging cases.

### 3.1. DC-RPN

While the conventional Mask R-CNN considers rectangular RoIs, an iris region is approximately specified by two non-concentric circles. Hence, we propose to generate a set of double-circle region proposals by a DC-RPN to capture the RoIs associated with iris segmentation. DC-RPN can be conceptually considered as a sliding-window detector. It uses a small network sliding over feature maps generated by the CNN backbone network to perform iris/non-iris classification and double-circle regression. The classification and regression target are defined with respect to a set of reference double-circles referred to as the anchors.

The architecture of DC-RPN is illustrated in Figure 2. It is implemented with 512 3×3 convolutional layers with ReLU activation, followed by two sibling 1×1 convolutional layers for classification and double-circle regression respectively.

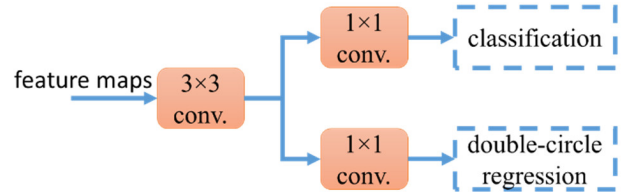

Figure 2: The network architecture is shown for DC-RPN.

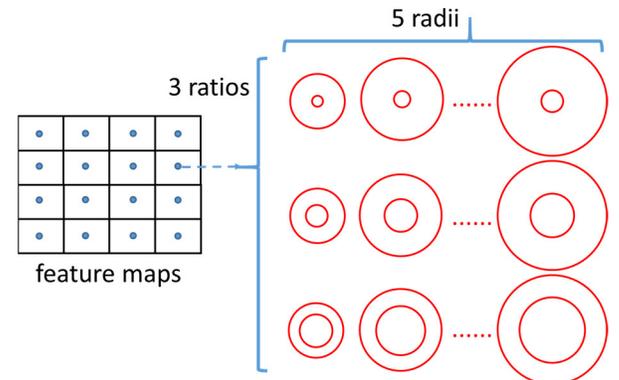

Figure 3: Fifteen double-circle anchors are generated with five different radii of the iris circle and three different ratios between the radii of iris and pupil circles.

**Anchors**: An anchor is a double-circle in DC-RPN. A set of double-circle region proposals are generated by regressing offsets from the corresponding anchors while sliding the DC-RPN over the feature maps. Each double-circle anchor is parameterized by a 6-tuple $(x_I^A, y_I^A, x_P^A, y_P^A, r_I^A, r_P^A)$, where the first four elements specify the center coordinates of the iris and pupil circles and the last two elements specify the radii of these two circles.

To appropriately capture the iris regions with different sizes, we use two parameters to control the size of each anchor: (i) the radius of the iris circle, and (ii) the ratio between the radii of iris and pupil circles, as shown in Figure 3. At each location on a feature map, we generate fifteen anchors by using five database-specific radii and three different ratios {0.1, 0.2, 0.5}. For a feature map with $W \times H$ pixels, there are $W \times H \times 15$ anchors in total.

**Classification**: In Figure 2, the convolutional layer for classification predicts two scores estimating the probabilities for each anchor to be iris and non-iris, respectively. It is implemented with a two-class softmax layer. For the fifteen anchors generated at each location on a feature map, it outputs $2 \times 15 = 30$ scores as the classification outcome.

**Double-circle regression**: The goal of double-circle regression is to learn a transformation that maps a proposed double-circle anchor $P$ to the ground-truth double-circle $G$. Denote the proposal $P$ as $(x_I^A, y_I^A, x_P^A, y_P^A, r_I^A, r_P^A)$ and the ground truth $G$ as $(x_I^G, y_I^G, x_P^G, y_P^G, r_I^G, r_P^G)$. We parameterize the transformation by six parameters $(t_x^I, t_y^I, t_x^P, t_y^P, t_r^I, t_r^P)$, similar to [21]. The first four parameters specify the center offsets for the iris and pupil circles, and the last two parameters specify the scaling factors for their radii:

$$x_I^G = x_I^A + t_x^I r_I^A \quad y_I^G = y_I^A + t_y^I r_I^A \quad (1)$$

$$x_P^G = x_P^A + t_x^P r_P^A \quad y_P^G = y_P^A + t_y^P r_P^A \quad (2)$$

$$r_I^G = r_I^A \cdot \exp(t_r^I) \quad r_P^G = r_P^A \cdot \exp(t_r^P). \quad (3)$$

For each anchor, the aforementioned six parameters are predicted by the convolutional layer for regression in Figure 2. Given the fifteen anchors generated at each location on a feature map, the regression layer outputs $6 \times 15 = 90$ parameter values in total.

**Learning proposals**: Given a dataset labelled with the ground truth of iris and pupil circles, we assign the training label to each anchor based on its Intersection-over-Union (IoU) ratio over the ground truth. When calculating the IoU of an anchor, we first calculate two separate IoUs for its iris and pupil circles, respectively. The average of these two IoUs is defined as the final IoU for the anchor. To reduce the computational cost, we further calculate the circumscribed squares for both circles and estimate the IoUs based on these two squares, instead of two circles. During the training process, an anchor is assigned with a positive label if its IoU is greater than 0.7 and a negative label if its IoU is less than 0.3. Anchors without positive or negative labels are not used for training.

Similar to Mask R-CNN, we adopt a multi-task loss function for training:

$$L_{RPN}(\mathbf{w}) = L_{CLS}(\mathbf{w}) + \lambda \cdot L_{REG}(\mathbf{w}), \quad (4)$$

where $L_{CLS}$ denotes the softmax loss for the classification layer, $L_{REG}$ represents the smooth-$L_1$ loss for the regression layer defined on positive anchors only, $\mathbf{w}$ contains the unknown network weights to be trained, and $\lambda$ is a user-defined hyper-parameter balancing $L_{CLS}$ and $L_{REG}$.

### 3.2. RoI Normalization

After the input image is processed by DC-RPN, we are left with a number of double-circle region proposals of different sizes. In practice, the image size of iris may vary, because it is captured from different people, with different distances, at different angles and/or under different lighting conditions. On the other hand, the subsequent neural networks for iris segmentation and recognition are often designed for fixed-size feature maps. Therefore, it is crucial to extract a fixed-size representation for each double-circle region to interface with the subsequent segmentation and recognition branches.

Conventionally, RoI pooling [22] is a standard operation used for object detection to extract a fixed-size feature vector from variable-size feature maps corresponding to different region proposals. It divides a given RoI into $K \times K$ bins by a regular grid and then max-pools the values in each RoI bin to form a $K \times K$ grid cell. The RoIAlign method used by Mask R-CNN [14] is an improved pooling operation that uses a bilinear interpolation to compute the features at four regularly sampled locations in each RoI bin.

The aforementioned RoI pooling, however, cannot be directly applied to a double-circle region, because it relies on a regular sampling grid to perform the pooling operation. It, in turn, motivates us to develop a radically new RoI normalization scheme to generate a fixed-size representation for each double-circle region.

Towards this goal, we borrow the idea of the rubber-sheet model [3], which is widely used for iris normalization. The proposed RoI normalization is simple: it unwraps each double-circle region into a rectangular region by remapping the original Cartesian coordinates to the dimensionless polar coordinates, as shown in Figure 4. More details about the rubber-sheet model can be found in [3] and they are not included in this paper due to page limit.

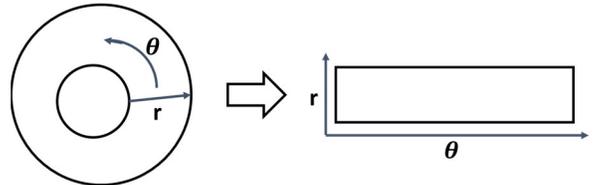

Figure 4: The proposed RoI normalization unwraps each double-circle region into a rectangular region using the rubber-sheet model.

The normalized feature maps generated by RoI normalization are fed into the DC-CRN and the mask network for subsequent processing. The details about these two subsequent networks will be discussed in the following sub-sections.

### 3.3. DC-CRN

The DC-RPN in Section 3.1 proposes, classifies and locates a number of double-circle regions. Since the accuracy of DC-RPN is often limited in practice, we further refine the outcomes of classification and regression by DC-CRN based on the normalized RoI features in order to improve the accuracy. Such a refinement operation is similarly adopted by Mask R-CNN [14] in the literature.

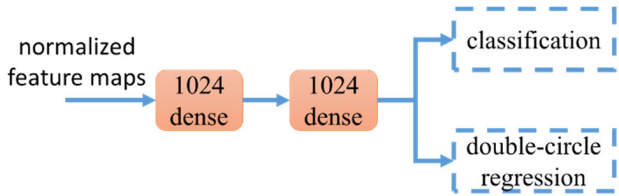

Figure 5: The network architecture is shown for DC-CRN.

The architecture of DC-CRN is illustrated in Figure 5. It is implemented with two 1,024-dimensional FC layers, each using ReLU activation and batch normalization, followed by two sibling FC layers for iris/non-iris classification and double-circle regression respectively.

The FC layer for classification predicts two scores estimating the probabilities for each proposed RoI to be iris and non-iris, respectively. It is implemented with a two-class softmax layer, similar to Mask R-CNN.

Unlike the conventional Mask R-CNN that focuses on rectangular RoIs, our proposed FC layer for regression aims to capture two non-concentric circles, corresponding to the iris and pupil circles respectively. By adopting the idea of double-circle regression in Section 3.1, the FC layer for regression in Figure 5 outputs a 6-tuple ($t_x^I, t_y^I, t_x^P, t_y^P, t_r^I, t_r^P$), encoding the final refinement on locations and sizes of these two circles. The center coordinates and radii of both circles are calculated from ($t_x^I, t_y^I, t_x^P, t_y^P, t_r^I, t_r^P$) by following the mathematic equations in (1)-(3).

To train DC-CRN, each normalized RoI is configured with a fixed spatial extent of 7×7. The RoI is considered to be positive if its IoU is no less than 0.5 and negative otherwise. The multi-task loss in (4) is used to jointly train the classification and regression layers based on labeled RoIs.

### 3.4. Mask Network

The mask network is implemented with an FCN that takes the normalized RoI features as its input and generates the corresponding normalized binary masks. Note that our proposed approach for mask generation is substantially different from most conventional FCN-based methods where a binary mask is generated for the entire iris image, instead of the RoI.

In this paper, we adopt a radically different approach for mask generation due to two important reasons. Firstly, if a full iris image is used, several non-iris regions, such as hairs and eyebrows, are similar to iris and they may be classified as iris by mistake. Secondly, the conventional approaches must perform an additional mask normalization step to provide the required normalized mask for subsequent iris recognition, unnecessarily increasing the computational complexity.

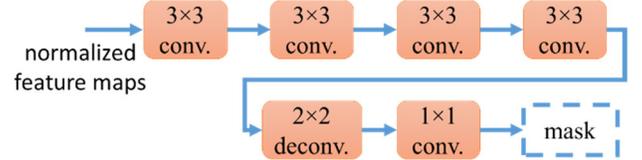

Figure 6: The network architecture is shown for the mask network.

The architecture of our mask network is illustrated in Figure 6. Similar to Mask R-CNN, it is composed of four 3×3 convolutional layers followed by 2×2 transposed convolutional layers with stride 2, and a final 1×1 convolutional layer outputting the final binary mask. The output resolution of the mask network is set to 32×64. The mask network is trained by minimizing the average binary cross-entropy loss.

## 4. Iris Recognition

The recognition accuracy is always the first concern for iris recognition systems. In order to demonstrate that our proposed iris segmentation can enhance the overall recognition accuracy, we further combine Iris R-CNN with a recognition network to form a complete iris recognition system.

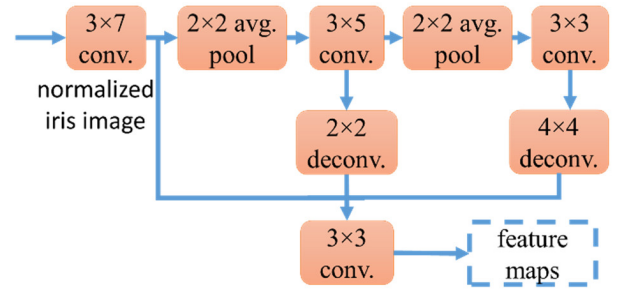

Figure 7: The network architecture is shown for the recognition network.

The architecture of the recognition network is summarized in Figure 7. It adopts FeatNet, an FCN specifically designed for iris recognition in [18], as the encoding network. The network is composed of a number of convolutional layers, activation layers and pooling layers, which gradually reduce the sizes of feature maps. Next, the feature maps at different scales are up-sampled by deconvolutions to match the original size of normalized iris image. As such, the network is able to capture both the local and global information for iris recognition. An extended triplet loss function incorporating bit-shift and mask is used to train the network to extract discriminative spatial iris features. More details about the recognition network and the extended triplet loss can be found in [18].

During evaluation, a similarity score is computed based on the fractional Euclidean distance between two masked feature maps for matching purpose:

$$D(A,B) = \frac{1}{|\Phi|} \cdot \sum_{(x,y) \in \Phi} \left( f_{x,y}^A - f_{x,y}^B \right)^2, \quad (5)$$

where $A$ and $B$ stand for the feature maps, $f_{x,y}^A$ and $f_{x,y}^B$ are the $(x,y)$-th pixels of these two feature maps respectively, $\Phi$ is a set defined by:

$$\Phi = \left\{ (x,y) \big| m_{x,y}^A \neq 0 \text{ and } m_{x,y}^B \neq 0 \right\}, \quad (6)$$

and $|\bullet|$ represents the cardinality of a set (i.e., the total number of elements in the set). In (6), $m_{x,y}^A$ and $m_{x,y}^B$ denote the $(x,y)$-th pixels of the binary masks for $A$ and $B$, respectively. A spatial location is non-iris, if the corresponding mask value is zero.

## 5. Experimental Results

In this section, we present the experimental results based on two challenging databases, (i) NICE-II [23] and (ii) MICHE [24], to demonstrate the superior accuracy for the proposed iris segmentation and recognition framework. These two databases are chosen for our experiments because their iris images are acquired under non-cooperative environment with visible light and, therefore, cannot be easily segmented and recognized by other conventional methods.

### 5.1. Segmentation Accuracy

Table 1: Summary of two databases (i.e., NICE-II [23] and MICHE [24]) for iris segmentation.

|  |  |  | Number of subjects | Number of images |
|---|---|---|---|---|
| NICE-II | | Training | 171 | 1,000 |
| | | Testing | 150 | 1,000 |
| MICHE | GS4 | Training | 63 | 215 |
| | | Testing | 75 | 634 |
| | IP5 | Training | 62 | 305 |
| | | Testing | 75 | 631 |
| | GT2 | Training | 41 | 116 |
| | | Testing | 75 | 316 |

**NICE-II**: It is a subset of the UBIRIS.v2 database [25]. NICE-II contains iris images captured under heterogeneous lighting conditions without infrared illumination, thereby leading to highly degraded image quality. It contains two non-overlapping subsets: (i) the training set with 1,000 images from 171 subjects, and (ii) the testing set with 1,000 images from 150 subjects, as shown in Table 1. The dataset provides labelled masks as the ground truth. We manually annotate the iris and pupil circles based on the labelled masks for the training set.

**MICHE**: Its iris images are acquired by three different mobile devices, including (i) iPhone5 (IP5), (ii) Samsung Galaxy S4 (GS4), and (iii) Samsung Galaxy Tab2 (GT2), with two different acquisition modes (i.e., indoor vs. outdoor) under non-cooperative conditions without infrared illumination. MICHE covers 75 different subjects in total, with 1,297 images from GS4, 1,262 images from IP5, and 632 images from GT2. In our experiment, we use the indoor images for training and the outdoor images for testing. For this database, since the labelled masks are not provided, we first use the algorithm from [8] to generate the labelled segmentation results, and then manually remove the iris images that are incorrectly labelled by [8]. Here, the algorithm of [8] is chosen because it has been demonstrated in the literature to offer superior segmentation accuracy for MICHE. Given the aforementioned setup, 215 out of 663 images are selected as the training set for GS4, 305 out of 631 images are selected as the training set for IP5 and 116 out of 316 images are selected as the training set for GT2, as shown in Table 1.

Figure 8 shows several representative iris images from NICE-II and MICHE. Note that these images carry substantial artifacts, as they are acquired under various non-cooperative conditions. Consequently, accurate iris segmentation and recognition are extremely challenging in these cases.

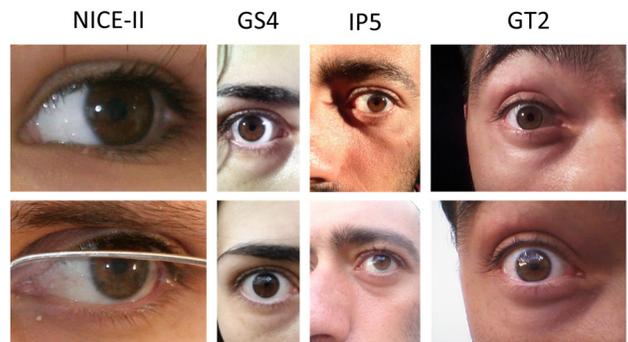

Figure 8: Iris images with substantial artifacts are shown for the two databases: (i) NICE-II and (ii) MICHE (GS4, IP5 and GT2).

We compare Iris R-CNN against a state-of-the-art iris segmentation approach based on Total Variation Model (TVM) [9]. TVM has been demonstrated as one of the promising methods in the literature. It adopts a novel formulation to robustly suppress noisy texture pixels for accurate iris segmentation. TVM generates the iris and pupil circles, as well as the corresponding segmentation mask. Its source code and parameter setups have been released to the public and they are used in our experiment. Note that we do not compare Iris R-CNN against other deep learning approaches in the literature for iris segmentation, because these deep learning methods do not explicitly detect the iris and pupil circles and thus do not explicitly provide a normalized iris mask.

We implement the proposed Iris R-CNN architecture based on [28]. The network is initialized by using a pre-trained model on the Microsoft COCO dataset [29], and is then trained for 60 epochs where the learning rate is reduced by 10× after 30 epochs. The mini-batch stochastic gradient descent with momentum of 0.9 and weight decay of 0.0001 is used as the optimizer. The resolution of the normalized iris image and mask is set to 64×512 for all experiments.

We introduce two quality metrics to quantitatively evaluate the accuracy of iris segmentation. First, we compute the IoU between the detected double-circle region and the labelled mask:

$$IoU_{SEG} = \frac{|\Omega_E \cap \Gamma_L|}{|\Omega_E \cup \Gamma_L|}, \quad (7)$$

where $\Omega_E$ and $\Gamma_L$ are the sets of pixels for the detected double-circle region and the labelled mask, respectively. A greater $IoU_{SEG}$ value corresponds to a larger overlap between $\Omega_E$ and $\Gamma_L$, thereby implying higher segmentation accuracy.

Second, we report the average mask error:

$$Err_{SEG} = \frac{1}{W \cdot H} \cdot \sum_{x=1}^{W} \sum_{y=1}^{H} \left( m_{x,y}^L \oplus m_{x,y}^E \right), \quad (8)$$

where $m_{x,y}^L$ and $m_{x,y}^E$ denote the $(x,y)$-th pixels of the labelled and estimated masks respectively, $W$ and $H$ are the weight and height of the normalized iris image respectively, and $\oplus$ represents the operation of Exclusive-OR (XOR) for two binary values. Note that the average mask error $Err_{SEG}$ is calculated after mask normalization, where the normalization step is based on the predicted iris and pupil circles.

Table 2: Iris segmentation accuracy on NICE-II.

|  | $IoU_{SEG}$ (%) | $Err_{SEG}$ (%) |
| --- | --- | --- |
| TVM [9] | 96.6 ± 4.0 | 10.1 ± 7.3 |
| Iris R-CNN | 97.1 ± 2.0 | 8.9 ± 2.8 |

Table 2 compares the accuracy of iris segmentation for TVM [9] and Iris R-CNN based on NICE-II. As the labelled masks are not provided for MICHE, we cannot report the segmentation accuracy for MICHE. As shown in Table 2, Iris R-CNN achieves superior accuracy over TVM in our experiment. Note that both the mean and variance are improved by Iris R-CNN for $IoU_{SEG}$ and $Err_{SEG}$. The significant impact of such an improvement in segmentation accuracy will be further demonstrated by the final recognition accuracy in the next sub-section.

## 5.2. Recognition Accuracy

The two databases, NICE-II and MICHE, are again used to compare the recognition accuracy for a variety of methods, as summarized in Table 3. For NICE-II, we follow the same practice described in the previous sub-section to form the training and testing sets. As a result, the testing set generates 4,634 genuine pairs and 494,866 imposter pairs for accuracy evaluation. For MICHE, we use all indoor images for training and all outdoor images for testing. With this setup, the testing set of GS4 generates 2,510 genuine pairs and 198,151 imposter pairs, the testing set of IP5 produces 2,366 genuine pairs and 196,399 imposter pairs, and the testing set of GT2 results in 534 genuine pairs and 49,236 imposter pairs.

Table 3: Summary of two databases (i.e., NICE-II [23] and MICHE [24]) for iris recognition.

|  |  |  | Number of subjects | Number of images |
| --- | --- | --- | --- | --- |
| NICE-II |  | Training | 171 | 1,000 |
|  |  | Testing | 150 | 1,000 |
| MICHE | GS4 | Training | 75 | 663 |
|  |  | Testing | 75 | 634 |
|  | IP5 | Training | 75 | 631 |
|  |  | Testing | 75 | 631 |
|  | GT2 | Training | 75 | 316 |
|  |  | Testing | 75 | 316 |

We compare Iris R-CNN against three conventional approaches, both with and without using deep learning:

**ICCV 2015 [9]**: It relies on the TVM method [9] for accurate iris segmentation. Next, a 1-D log-Gabor filter is adopted for feature extraction. Its source code for iris segmentation has been released to the public and it is used in our experiment. On the other hand, the 1-D log-Gabor filter in [26] is used for iris recognition.

**ICIP 2016 [17]**: A publically available system, Osiris v4.1 [27], is used for iris segmentation, and two CNN networks are supervised by a softmax loss function for iris recognition. The approach is referred to as DeepIrisNet. In our experiment, we carefully follow [17] to re-implement DeepIrisNet for testing and comparison purposes.

**ICCV 2017 [18]**: It relies on the TVM method [9] for iris segmentation. Next, two FCN networks are trained for iris recognition. The first FCN network, referred to as FeatNet, takes a normalized iris image as its input and generates the required feature maps for matching. The second FCN, referred to as MaskNet, takes a normalized iris image and generates the binary mask. In our experiment, we carefully implement both FCN networks according to [18] in TensorFlow. Note that the aforementioned flow is different from our proposed Iris R-CNN in iris segmentation, while they adopt the same approach for iris recognition.

For matching any two iris images, we use the fractional Euclidean distance defined in (5). If the segmentation algorithm fails to generate any meaningful results and the set Φ in (5) is empty, we assign the corresponding distance value to be infinite. The accuracy of iris recognition is evaluated by using both the Equal Error Rate (EER) and the Receiver Operating Characteristic (ROC) curve, as shown in Table 4 and Figure 9.

Table 4: Iris recognition accuracy on NICE-II and MICHE. (The symbol "–" means that no valid EER can be reported due to large segmentation error.)

|  | EER (%) | | | |
| --- | --- | --- | --- | --- |
|  | NICE-II | GS4 | IP5 | GT2 |
| ICCV 2015 [9] | 24.0 | 49.1 | 45.0 | 30.5 |
| ICIP 2016 [17] | 48.9 | – | – | – |
| ICCV 2017 [18] | 23.2 | 49.0 | 47.0 | 31.1 |
| Iris R-CNN | 13.5 | 32.4 | 25.0 | 16.7 |

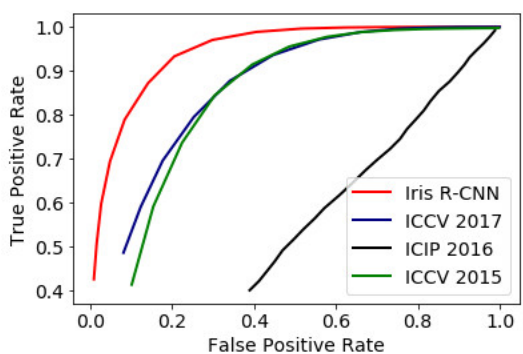

(a) NICE-II

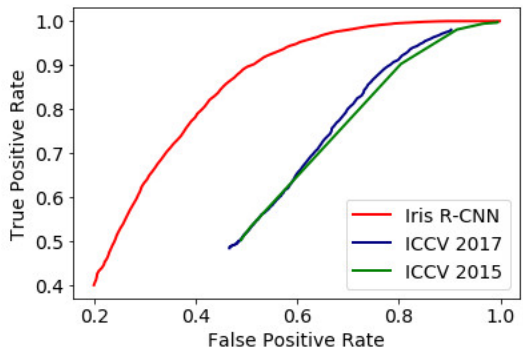

(b) GS4

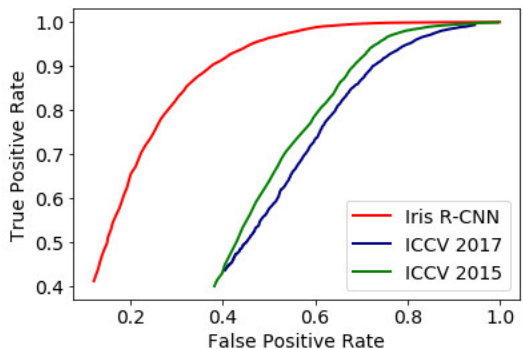

(c) IP5

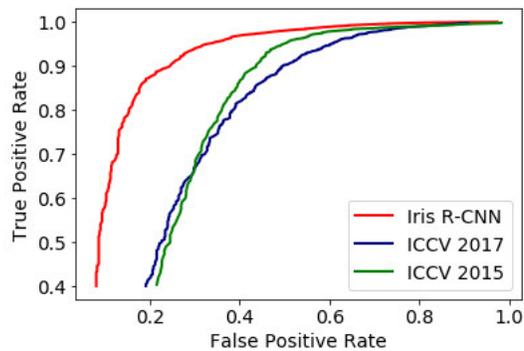

(d) GT2

Figure 9: ROC curves are shown for iris recognition based on two databases: (i) NICE-II and (ii) MICHE (GS4, IP5 and GT2). No meaningful ROC curve can be generated for MICHE by ICIP 2016 due to large segmentation error.

Our proposed Iris R-CNN substantially improves the recognition accuracy over other state-of-the-art methods. These results strongly suggest that the accuracy of iris recognition is highly dependent on the quality of iris segmentation. While iris segmentation has been extensively studied in the literature, it remains a grand challenge in non-cooperative environment. By developing an enhanced technique for iris segmentation, we have demonstrated significantly improved accuracy of iris recognition for two challenging databases: NICE-II and MICHE.

## 6. Conclusions

In this paper, we present a deep learning framework, referred to as Iris R-CNN, to offer superior accuracy for iris segmentation. The proposed framework is derived from Mask R-CNN, and several novel techniques are proposed to carefully explore the unique characteristics of iris, including (i) DC-RPN, (ii) DC-CRN, and (iii) RoI normalization. Based on two challenging databases (i.e., NICE-II and MICHE), superior accuracy in both iris segmentation and recognition has been demonstrated for Iris R-CNN against other state-of-the-art methods.